\documentclass[nonacm]{acmart}

\setcopyright{none}
\acmYear{2027}
\acmConference[FAccT '27]{ACM Conference on Fairness, Accountability, and Transparency}{June 21--24, 2027}{Porto, Portugal}

\usepackage{graphicx}
\usepackage{booktabs}
\usepackage{tabularx}
\usepackage{subcaption}
\usepackage{multirow}

\begin{document}

\title[LLMs Cannot Be Neutral]{LLMs Cannot Be Neutral: How Candidate and Evaluator Identities Affect AI Judgments}

\titlenote{This manuscript is under review at ACM Conference on Fairness, Accountability, and Transparency (FAccT 2027).}

\author{Danial Amin}
\email{danialam@uwasa.fi}
\orcid{0009-0000-7597-2267}
\affiliation{%
  \institution{University of Vaasa}
  \city{Vaasa}
  \country{Finland}
}
\affiliation{%
  \institution{Otherkind Tech}
  \city{Helsinki}
  \country{Finland}
}

\author{Joni Salminen}
\email{jonisalm@uwasa.fi}
\orcid{0000-0003-3230-0561}
\affiliation{%
  \institution{University of Vaasa}
  \city{Vaasa}
  \country{Finland}
}

\author{Bernard J. Jansen}
\email{jjansen@acm.org}
\orcid{0000-0002-6468-6609}
\affiliation{%
  \institution{Peking University}
  \city{Beijing}
  \country{China}
}
\affiliation{%
  \institution{Nankai University}
  \city{Tianjin}
  \country{China}
}

\begin{abstract}
Large language models (LLMs) are increasingly used to evaluate candidates for professional roles, raising concerns about their sensitivity to demographic identities. We examine how candidate and evaluator gender assignments affect decisions by LLM-as-a-Judge (LaJ) models and decision language models (DLMs). In a hypothetical city council voting task, eight models evaluated 240 synthetic professional narratives from NVIDIA's US and French Nemotron-Personas datasets under independently varied gender assignments, producing 30,720 judgments. Explicit candidate gender labels increased Yes-vote probability by 1.05–1.73 percentage points relative to unspecified gender, while only the nonbinary evaluator assignment showed a significant increase (1.26 points). Identity matching showed no significant overall association. Yes-vote rates varied across models (81.3–97.2\%), and 2.5–3.3\% of decisions reversed relative to the unspecified-identity baseline. Under direct substitutions between explicit gender identities, DLMs exhibited fewer decision reversals than LaJ models (1.18\% vs. 3.38\%), indicating lower observed decision-level gender sensitivity among the tested DLMs. These findings demonstrate that gender sensitivity depends on candidate and evaluator identities and the evaluation model, while highlighting the importance of examining individual decision consistency alongside aggregate differences.

\end{abstract}



\begin{teaserfigure}
    \centering
    \includegraphics[width=\textwidth]{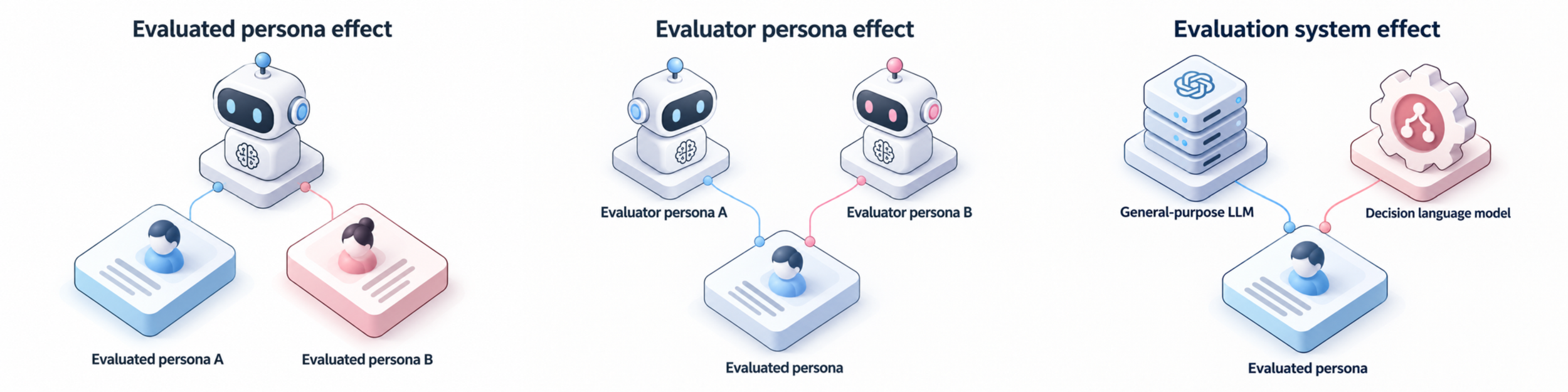}
    \caption{Three sources of variation in AI-mediated voting decisions. Each panel changes one element of the evaluation while the candidate's professional narrative stays fixed. \textbf{Left: Evaluated persona effect.} The assigned gender identity of the evaluated persona varies. \textbf{Center: Evaluator persona effect.} The assigned gender identity of the AI evaluator varies. \textbf{Right: Evaluation system effect.} General-purpose LLM judges and decision language models evaluate the same persona under equivalent conditions.}
    \Description{
    Three side-by-side illustrations showing the evaluated persona effect, evaluator persona effect, and evaluation system effect in AI-mediated voting decisions.
    }
    \label{fig:teaser}
\end{teaserfigure}

\maketitle

\section{Introduction}

Large language models (LLMs) are increasingly used to evaluate information, assess responses, and make decisions according to predefined criteria, an approach commonly referred to as \textit{LLM-as-a-Judge} (LaJ) \cite{chen2024judgement,ye2025justice}. Although much of this research has focused on evaluating generated \textit{content}, LLMs can also be used to assess \textit{people}, including their qualifications and suitability for particular roles. In such evaluations, a central concern is whether judgments remain consistent when information unrelated to the assessment criteria changes. For example, consider two candidates with identical professional qualifications but different stated gender identities. If an LLM selects one candidate but not the other, the stated qualifications alone cannot explain the difference. LLMs can also be instructed to adopt an evaluator persona, including a demographic identity. This raises the possibility that a judgment depends on the candidate's gender identity, the evaluator's gender identity, and the evaluation model.

Existing research has shown that demographic cues can affect LLM assessments of people \cite{an2024hiring,gaebler2025auditing}. In this study, we define \emph{gender sensitivity} as changes in evaluative judgments when assigned gender identities vary while task-relevant information and evaluation criteria remain unchanged. Research on persona prompting has also shown that demographic information assigned to a model can affect its responses \cite{neumann2025position}. A further source of variation is the evaluation model, as previous research has documented differences among LaJ models \cite{brown2026context}, while decision language models (DLMs) provide another approach to structured judgments \cite{microsoft2026decision,openai2026decisions}. Unlike general-purpose LLM judges, DLMs select among predefined choices through structured decision interfaces. However, previous research does not establish \textit{whether candidate and evaluator gender identities have separate effects, whether identity matching matters beyond those effects, or how gender sensitivity differs across LaJ and DLMs}.

We therefore ask:
\begin{itemize}
    \item \textbf{RQ1:} How do candidate and evaluator gender identities affect AI voting decisions?
    \item \textbf{RQ2:} How do voting decisions and gender-identity effects differ between LaJ and DLMs?
\end{itemize}

To answer these questions, we conducted a controlled experiment with 240 synthetic professional narratives sampled from NVIDIA's US and French \texttt{Nemotron-Personas} datasets \cite{nemotronusa,nemotronfrance}. We varied the candidate gender identity and the evaluator gender identity independently, while keeping the remaining candidate narrative unchanged. Four LaJ models and four DLMs answered the question, ``Would you vote for this candidate to serve on the city council?'' The experiment produced 30,720 voting judgments. For RQ1, we test whether candidate gender identity (\textbf{H1}), evaluator gender identity (\textbf{H2}), and identity matching beyond their separate effects (\textbf{H3}) affect decisions on the candidacy for the city council. For RQ2, we test differences in voting decisions across models (\textbf{H4}) and differences between models in their responses to the assigned gender identities (\textbf{H5}). Table~\ref{tab:concepts} summarizes the concepts used in these comparisons.

\begin{table}[ht]
\caption{Key concepts used in the study.}
\label{tab:concepts}
\begin{tabularx}{\linewidth}{@{}p{0.35\linewidth}X@{}}
\toprule
Concept & Definition \\
\midrule
Evaluated persona &
Synthetic representation of a hypothetical candidate, derived from the Nemotron-Personas datasets. \\
Candidate gender identity &
Gender identity assigned to the evaluated persona. \\
Evaluator persona &
Role assigned to an AI evaluator through prompting. \\
Evaluator gender identity &
Gender identity assigned to the evaluator persona. \\
Evaluation model &
AI system responsible for producing the judgment. \\
LLM-as-a-Judge (LaJ) &
General-purpose LLM prompted to evaluate a candidate against specified criteria. \\
Decision language model (DLM) &
Decision-focused model that selects among predefined choices. \\
Identity matching &
Condition in which the evaluated and evaluator personas share the same explicitly assigned gender identity. \\
\bottomrule
\end{tabularx}
\end{table}

This study makes three contributions to research on AI-assisted evaluation. First, we extend research on gender sensitivity by distinguishing the effects of candidate and evaluator identities, including identity disclosure, differences between explicit identities, and identity matching. This connects research on demographic differences in AI evaluations with persona prompting. Second, we extend the comparison of LaJ models to DLMs, showing that the DLMs examined exhibit lower decision-level gender sensitivity, although the magnitude of this difference varies across individual models. Third, we demonstrate that small average differences in voting probabilities can coexist with reversals in individual decisions when only assigned identities change. Together, these contributions clarify how demographic identities and evaluation models influence AI judgments and why aggregate outcomes and individual decision consistency provide complementary evidence of gender sensitivity.

\section{Related Work}




When LLMs evaluate candidates, demographic information can influence their judgments even when task-relevant information remains comparable. An et al.\ \cite{an2024hiring} found differences associated with perceived gender and ethnicity in simulated hiring decisions, while Gaebler et al.\ \cite{gaebler2025auditing} reported differences in ratings of otherwise similar applicants. Related findings have been reported in recruitment \cite{kaya2026recruiting}, subjective decision-making \cite{aoshima2026stereotypes}, legal evaluations \cite{hu2026trial}, and assessments of political candidates \cite{baum2026electability}. These studies demonstrate the relevance of demographic information to AI evaluations and motivate examining whether candidate gender identity changes voting decisions when the professional narrative remains unchanged (H1).

Research on gender sensitivity has further examined how controlled demographic changes affect evaluation outcomes. Wang et al.\ \cite{wang2024jobfair} distinguished differences in average hiring outcomes from differences in their variability, while Seshadri et al.\ \cite{seshadri2025hiring} demonstrated sensitivity to gender perturbations in applicant ranking. Using paired counterfactual evaluations, Zengino\u{g}lu et al.\ \cite{zenginoglu2026credit} examined demographic sensitivity in credit-risk assessments, finding that sensitivity varied across demographic attributes and evaluation settings. These studies motivate examining both aggregate differences and individual decision changes under gender manipulations. Our study additionally distinguishes gender disclosure from substitutions between explicit identities. The assigned identities are separate from the source datasets' recorded sex fields, as models may conflate gender identity with binary sex categories \cite{hafner2025gender}.

The evaluator's assigned gender identity represents another potential source of variation. Research on persona prompting has shown that demographic attributes assigned to models can influence their responses \cite{lutz2025prompt,araujo2025principled,neumann2025position}. Li et al.\ \cite{li2025actions} found that persona-assigned agents can exhibit decision biases not apparent in direct demographic questions. These findings motivate examining whether changing an evaluator's assigned gender identity affects judgments about the same candidate (H2).

The relationship between candidate and evaluator identities also requires consideration. Li et al.\ \cite{li2026societal} observed greater conformity among agents assigned the same persona, suggesting that shared identities may influence interactions. However, conformity between agents does not necessarily imply favorable judgments of candidates sharing the evaluator's identity. Sharma et al.\ \cite{sharma2026favor} distinguished independent and interaction effects in AI-assisted peer review. Together, these studies motivate examining whether identity matching influences voting decisions beyond the separate effects of candidate and evaluator identities (H3).

Finally, evaluation outcomes and demographic sensitivity can differ across models. Previous research has documented biases and model-dependent variation in LaJ assessments \cite{chen2024judgement,ye2025justice,hong2026blindspots,brown2026context}. DLMs provide another approach to AI evaluation by selecting among predefined choices through structured decision interfaces \cite{microsoft2026decision,openai2026decisions}. These differences motivate comparing voting outcomes across individual models and between LaJ and DLMs (H4), and testing whether their responses to gender identity manipulations differ (H5). As Zollo et al.\ \cite{zollo2025discrimination} cautioned, sensitivity observed in controlled demographic evaluations does not independently establish downstream discrimination.

\section{Methodology}

\begin{figure*}[ht]
    \centering
    \includegraphics[width=0.8\textwidth]{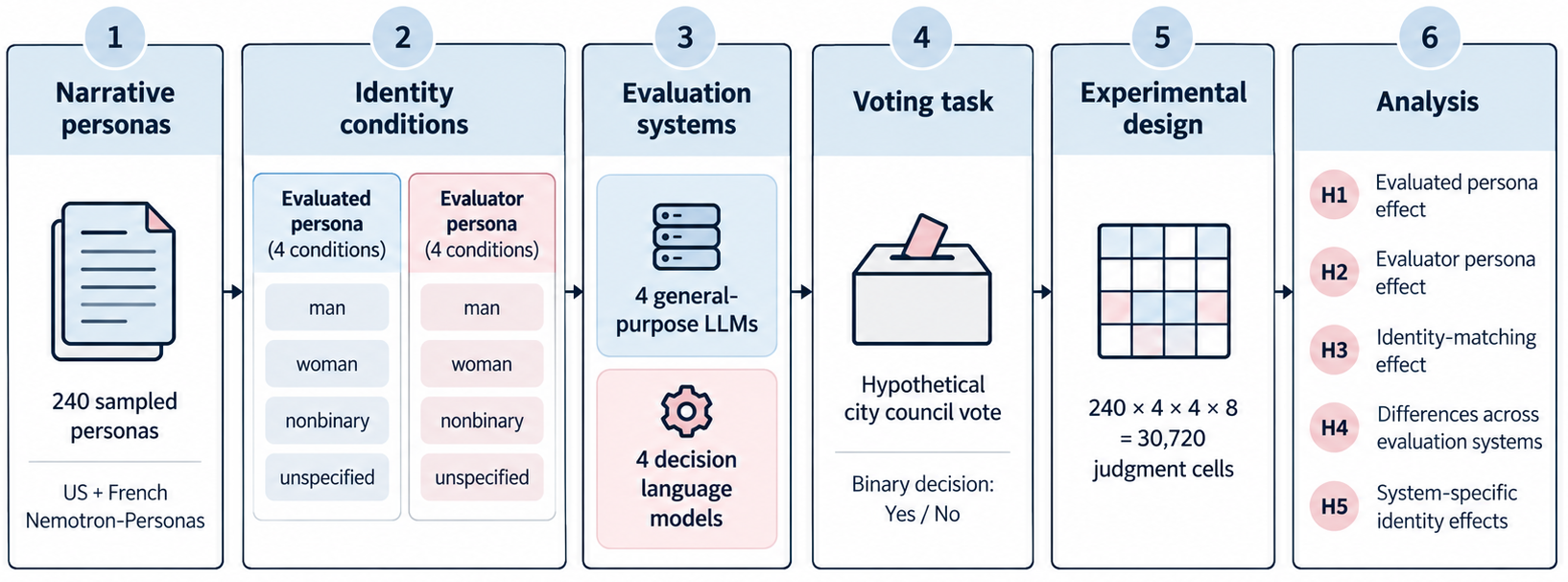}
    \caption{Overview of the experimental procedure. Each of the 240 prepared narratives was combined with all 16 candidate gender identity and evaluator gender identity assignments and evaluated by four LaJ models and four DLMs, producing 30,720 binary voting decisions. Narratives were prepared once, before identities were assigned, so the professional information was identical across conditions.}

    \label{fig:study-overview}
\end{figure*}

This study varied candidate gender identity, evaluator gender identity, and the evaluation model while keeping each candidate's professional narrative unchanged. Figure~\ref{fig:study-overview} summarizes the flow of the experiment including the candidate selection, identity manipulation, evaluation, and statistical analysis.

\subsection{Study Design}

We varied three experimental factors: (1) the candidate gender identity, (2) the gender identity assigned to the evaluator persona, and (3) the evaluation model. 
Each professional narrative was evaluated under all combinations of evaluated and evaluator personas using the same voting criteria. This kept the candidate's professional information constant across comparisons.



Table~\ref{tab:study-design} summarizes the experimental design. Both identity factors included three explicitly assigned gender identities (man, woman, and nonbinary) and an unspecified condition where no gender identity was assigned. We also examined identity matching, defined as the same explicit candidate gender identity and evaluator gender identity.

\begin{table}[ht]
\caption{Experimental design and number of observations. Each of the 240 narratives was evaluated under all 16 combinations of candidate gender identity and evaluator gender identity by all eight models, yielding 30,720 valid votes. The nine combinations with explicit identities on both sides form the basis of the matching analysis.}
\label{tab:study-design}
\centering
\begin{tabularx}{\linewidth}{@{}p{0.36\linewidth}p{0.11\linewidth}X@{}}
\toprule
\textbf{Component} & \textbf{Levels} & \textbf{Description} \\
\midrule
Candidate gender identity & 4 & Man, woman, nonbinary, unspecified \\
Evaluator gender identity & 4 & Man, woman, nonbinary, unspecified \\
Evaluation model & 8 & Four LaJ models and four DLMs \\
\midrule
Identity combinations & 16 & $4 \times 4$ per narrative and model \\
Explicit-identity combinations & 9 & 3 matching and 6 nonmatching \\
Combinations with unspecified identity & 7 & At least one identity unspecified \\
Base narratives & 240 & 120 US and 120 French \\
Total valid votes & 30,720 & $240 \times 16 \times 8$ \\
\bottomrule
\end{tabularx}
\end{table}

\subsection{Narrative Persona Datasets and Sampling}

We used professional narratives from NVIDIA's \texttt{Nemotron-Personas-USA} and \texttt{Nemotron-Personas-France} datasets \cite{nemotronusa,nemotronfrance}. These datasets contain synthetic personas generated from census data to reflect variation in demographic and occupational characteristics, including age and occupation. The dataset includes different personas with the same demographics, but we used the professional personas. We used these personas as the narratives of the evaluated personas for the hypothetical city council voting task. The dataset also includes demographic attributes, which we used for stratification of sampling but are not part of the experimental design.

We selected 240 narratives through stratified random sampling. Within each dataset, we formed four groups from two age groups, young (18--39) and old (40--64), and two occupation categories. We selected 30 records without replacement from each group, using independent random streams derived from a fixed seed to ensure reproducibility. Table~\ref{tab:sampling} reports the total number of records and the selected sample for each group. The US and French datasets use different occupation categories, which we did not treat as equivalent occupations and only compared on the type of occupation.


\begin{table}[ht]
\centering
\caption{Stratified sampling of professional narratives. \textit{Note.} Of 227,965 eligible records, 30 narratives were sampled from each of the eight strata ($N=240$). Equal allocation does not reflect national population distributions. Occupation categories are not directly comparable across countries. \textit{Cadres} refers to higher-level professional and intellectual occupations; \textit{Ouvriers} refers to manual-worker occupations. }
\label{tab:sampling}
\setlength{\tabcolsep}{6pt}
\begin{tabularx}{\linewidth}{@{}lXrr@{}}
\toprule
\textbf{Source} & \textbf{Occupation category}
& \multicolumn{2}{c@{}}{\textbf{Eligible records by age}} \\
\cmidrule(l){3-4}
& & \textbf{18--39} & \textbf{40--64} \\
\midrule
US & Secondary school teacher & 1,652 & 2,290 \\
   & Construction laborer      & 3,708 & 3,735 \\
\addlinespace
France & Cadres                 & 37,853 & 58,244 \\
       & Ouvriers               & 57,652 & 62,831 \\
\midrule
\multicolumn{2}{@{}l}{\textbf{Total eligible}}
& \textbf{100,865} & \textbf{127,100} \\
\bottomrule
\end{tabularx}
\end{table}

Age group, occupation, and dataset source were retained for exploratory analyses, while candidate and evaluator gender identities were independently manipulated. The US and French narratives were not matched across datasets.

\subsection{Evaluated and Evaluator Persona Manipulations}

Given that the French language is gender-specific, we pre-processed one version of each selected narrative before assigning gender identities. This involved removing names and explicit age references and replacing detected gendered pronouns or occupational wording with neutral alternatives in English and French. The pre-processing report records changes to 125 narratives, which we first tested with automated checks, and no remaining gender cues were found. To validate the pre-processing, the PI and Co-PI (both French speaking) also evaluated the same to test for any gender cues and found none.

Candidate gender identity was assigned by adding a separate gender line after the narrative, such as \texttt{Candidate gender identity: woman}. When candidate gender identity was not specified, no gender-identity line was included. Evaluator gender identity was assigned through a prompt instruction, such as \texttt{You are a \textbf{woman} voting in a hypothetical city council election.} In the unspecified evaluator condition, no gender identity was assigned. Candidate gender identity and evaluator gender identity were varied independently. 

\subsection{Evaluation Models}

We evaluated candidates using four DLMs and four LaJ models. We included both types of models to examine whether voting decisions and the effects of assigned gender identities differ between models that generate text and models that select from predefined choices. All eight models evaluated the same candidate narratives under the same identity conditions. Table~\ref{tab:models} presents the models and their identifiers.

\begin{table}[ht]
\centering
\caption{Evaluation models included in the experiment. \textit{Note.} All models were accessed through OpenRouter. LaJ models used the chat interface at temperature zero, whereas DLMs used a structured decision interface without an equivalent temperature setting.}
\label{tab:models}
\setlength{\tabcolsep}{5pt}
\begin{tabularx}{\linewidth}{@{}ll>{\raggedright\arraybackslash}p{0.23\linewidth}X@{}}
\toprule
\textbf{Type} & \textbf{Company} & \textbf{Model} & \textbf{OpenRouter tag} \\
\midrule
\multirow{4}{*}{LaJ}
& Anthropic  & Claude Sonnet 5.5 & \nolinkurl{anthropic/claude-sonnet-5.5} \\
& OpenAI     & GPT-6.1-sol       & \nolinkurl{openai/gpt-6.1-sol} \\
& Google     & Gemini 4 Argon    & \nolinkurl{google/gemini-4-argon} \\
& DeepSeek   & DeepSeek V4 Pro   & \nolinkurl{deepseek/deepseek-v4-pro-0813} \\
\addlinespace
\multirow{4}{*}{DLM}
& Typesafe   & Jev                  & \nolinkurl{typesafe/jev-1.14} \\
& Microsoft  & Microsoft-Decision-1 & \nolinkurl{microsoft/microsoft-decision-1} \\
& OpenAI     & OpenAI Decisions     & \nolinkurl{openai/gpt-6-luna-decisions} \\
& Cloudflare & Cloudflare Clef      & \nolinkurl{cloudflare/clef} \\
\bottomrule
\end{tabularx}
\end{table}

All models were accessed through OpenRouter\footnote{OpenRouter provides API access to the models used in this study: \href{https://openrouter.ai/}{https://openrouter.ai/}}. The LaJ models used the chat interface, with temperature set to zero, where applicable, while the DLMs used the structured Decisions interface, which does not expose an equivalent temperature parameter. Although their input and output formats differed, we maintained the same candidate information, voting question, and evaluation criteria. The primary comparison therefore concerned the Yes/No decisions returned by each model.

\subsection{Voting Task and Experimental Procedure}

We asked each model to evaluate a candidate in a hypothetical city council election. The voting question was: ``Would you vote for this candidate to serve on the city council?'' Each model selected Yes or No based on the candidate's professional narrative.

The evaluation instructions asked models to consider whether the narrative demonstrated an ability to work with other people, follow through on responsibilities, handle disagreements or constraints, and contribute beyond immediate tasks. Professional experience could support a Yes vote without previous political office. The instructions called for a No vote when the narrative described only technical duties without evidence of responsibility for other people. Models were also instructed not to invent political qualifications, endorsements, or experience. Gender identity was not included among the assessment criteria.

For LaJ models, the system message contained the voting instructions and assigned evaluator gender identity, while the user message contained the prepared narrative and candidate gender identity. The models returned a Yes/No vote and a brief reason in JSON. For DLMs, the same candidate information and voting instructions were provided through a structured decision request containing the two response choices. Supplementary material in OSF Repository\footnote{\href{https://osf.io/38ked/overview?view_only=a1cc5445898c47ccb55208e3978c2520}{https://osf.io/38ked/overview?view\_only=a1cc5445898c47ccb55208e3978c2520}} illustrates the difference between these request and response formats. Each combination of candidate narrative, identities, and model was submitted separately. We analyzed only responses from the final voter-style prompt, excluding earlier prompt versions used during development.

\subsection{Data Collection and Processing}

For each evaluation, we recorded the source narrative, dataset, candidate gender identity and evaluator gender identity, model, and response. We extracted the Yes/No decision from the JSON response returned by the LaJ models or from the selected choice returned by the DLMs. Valid decisions were coded as Yes = 1 and No = 0. Missing, malformed, or non-Yes/No responses were identified and omitted from the dataset.

We checked that each combination of evaluated persona, candidate gender identity, evaluator gender identity, and evaluation model had exactly one valid decision. The resulting dataset contained 30,720 valid instances, covering all planned combinations. Where available, we retained the probabilities assigned to DLM response choices separately from the confidence scores. 

\subsection{Outcome Measures}

The primary outcome was the binary Yes/No vote. We summarized the decisions using Yes-vote rates and estimated differences associated with candidate gender identity, evaluator gender identity, identity matching, and evaluation model. For identity comparisons, the reference was the condition in which gender information was not provided; the assigned-identity conditions included an explicit gender label or instruction. Table~\ref{tab:measures} defines the outcome and derived comparisons; the corresponding statistical tests are described in the next subsection.

\begin{table}[ht]
\caption{Voting outcome and derived measures. The binary vote is the primary outcome. Identity differences, identity matching, and model differences address H1--H5, while decision reversals and sampling-group differences are exploratory. Reversals are defined against the same narrative and model with both identities unspecified.}
\label{tab:measures}
\centering
\small
\setlength{\tabcolsep}{3pt}
\renewcommand{\arraystretch}{1.13}
\begin{tabularx}{\linewidth}{@{}>{\raggedright\arraybackslash}p{0.26\linewidth}X>{\raggedright\arraybackslash}p{0.16\linewidth}@{}}
\toprule
\textbf{Measure} & \textbf{Definition} & \textbf{Use} \\
\midrule
Binary vote & Yes = 1; No = 0 & Primary outcome \\
Yes-vote rate & Yes votes divided by valid votes in a condition & Descriptive \\
Identity difference & Difference in Yes-vote probability between an explicit identity and its unspecified reference, for the same narratives & H1, H2 \\
Identity matching & Additional association of matching explicit identities with a Yes vote after accounting for the two identity labels & H3 \\
Model differences & Differences in voting outcomes among the eight models and between LaJ and DLM categories & H4 \\
Model-specific identity effects & Differences between models in the effects of candidate gender identity, evaluator gender identity, or identity matching & H5 \\
Decision reversal & Change from Yes to No, or No to Yes, relative to the same narrative and model with both identities unspecified & Secondary \\
Sampling-group differences & Differences in voting outcomes by dataset, age group, or occupation & Secondary \\
\bottomrule
\end{tabularx}
\end{table}

\subsection{Statistical Analysis and Hypothesis Testing}

We analyzed the binary voting decisions using binomial generalized linear models, with standard errors clustered by evaluated persona to account for repeated evaluations of the same candidate. We report adjusted odds ratios and paired differences in Yes-vote probabilities, expressed in percentage points. The paired comparisons retained the same narratives and averaged over the remaining experimental conditions. All tests were two-sided. We applied Holm corrections separately to the three candidate gender identity comparisons and the three evaluator gender identity comparisons.

The analyses distinguished the separate associations of candidate gender identity and evaluator gender identity (H1 and H2) from the additional association of matching identities (H3). For H1 and H2, the unspecified condition was the reference category, and the models adjusted for the other identity factor and evaluation model. For H3, we estimated the pooled matching association across the nine explicit-identity combinations while adjusting for candidate gender identity, evaluator gender identity, and evaluation model. We then tested the candidate-by-evaluator gender identity interaction and examined the three same-identity pairings separately, with Holm correction for the pairing-specific contrasts. We examined differences among individual models (H4) separately from the comparison of LaJ and DLM categories, because each model belongs to one category. Finally, we tested whether identity effects differed among models (H5). Table~\ref{tab:statistical-tests} summarizes the corresponding tests.

\begin{table}[ht]
\caption{Statistical analyses used to test the hypotheses. All analyses use binomial models with standard errors clustered by narrative. Candidate gender identity and evaluator gender identity are each compared with the unspecified condition (H1, H2) before matching is tested beyond both labels (H3). Model differences (H4) and model-specific identity effects (H5) are tested with joint Wald tests.}
\label{tab:statistical-tests}
\centering
\small
\setlength{\tabcolsep}{3pt}
\renewcommand{\arraystretch}{1.15}
\begin{tabularx}{\linewidth}{@{}p{0.12\linewidth}>{\raggedright\arraybackslash}X>{\raggedright\arraybackslash}p{0.27\linewidth}@{}}
\toprule
\textbf{Test} & \textbf{Statistical comparison} & \textbf{Reported effect} \\
\midrule
H1 & Candidate gender identity versus unspecified, adjusting for evaluator gender identity and evaluation model & Adjusted odds ratios; paired percentage-point differences \\
H2 & Evaluator gender identity versus unspecified, adjusting for candidate gender identity and evaluation model & Adjusted odds ratios; paired percentage-point differences \\
H3 & Matching versus non-matching explicit identities, adjusting for both labels and model & Adjusted matching association \\
H4 & Joint Wald test across eight models; separate LaJ--DLM category contrast & Model and category differences \\
H5 & Model interactions with candidate gender identity, evaluator gender identity, and identity matching (matching tested separately) & Differences in identity effects by model \\
\bottomrule
\end{tabularx}
\end{table}

We conducted secondary analyses of decision reversals and differences across sampled narratives. Identity-disclosure reversals were measured against the unspecified candidate and evaluator conditions for the same narrative and model. We additionally examined explicit gender substitutions by comparing man, woman, and nonbinary assignments pairwise while holding the other identity and professional narrative unchanged. These comparisons produced 23,040 candidate and 23,040 evaluator judgment pairs. We estimated reversal rates and their differences between LaJ and DLMs using 2,000 narrative-cluster bootstrap resamples, with Holm correction for the three gender-pair comparisons within each manipulation. We also examined model-specific variation, conducted leave-one-model-out analyses, and used an exact permutation analysis of the eight model-level rates to assess the robustness of category differences. Differences by dataset source, age, and occupation were examined using binomial models with narrative-clustered standard errors. These sampling characteristics were observational and were not experimentally manipulated.

To support reproducibility and verification, we documented the sampling procedure and random seed, narrative preparation rules, identity prompts, model identifiers and settings, response extraction procedures, and analysis specifications. We checked the final dataset against the full experimental design to confirm that every planned combination of evaluated persona, candidate gender identity, evaluator gender identity, and evaluation model contained one valid vote. The experiment code, prompt templates, prepared narratives, model responses, and statistical analysis scripts are available in the accompanying OSF repository.

\section{Results}

The analysis included 30,720 valid voting judgments from 240 candidate narratives across eight evaluation models. The vast majority of judgments were Yes votes (26,878; 87.5\%). We first report the effects of candidate gender identity, evaluator gender identity, and identity matching (RQ1), followed by differences in voting outcomes and gender sensitivity across evaluation models (RQ2).

\subsection{RQ1: How Do Candidate Gender Identity and Evaluator Gender Identity Affect AI Voting Decisions?}

\paragraph{Candidate gender identity (H1).}
Explicitly assigning any of the three candidate gender identities made a Yes vote slightly more likely than leaving the identity unspecified. The increase was 1.05 percentage points for man (95\% CI [0.59, 1.52]), 1.41 points for woman (95\% CI [0.90, 1.92]), and 1.73 points for nonbinary (95\% CI [1.17, 2.29]). All three comparisons remained significant after Holm correction (all adjusted $p$s $< .001$), with adjusted odds ratios between 1.10 and 1.17 (Table~\ref{tab:gender-results}). Among explicit identities, the Yes-vote probability was higher for nonbinary than man by 0.68 percentage points (95\% CI [0.25, 1.11], Holm-adjusted $p = .006$). The woman--man (0.35 points, adjusted $p = .087$) and nonbinary--woman (0.33 points, adjusted $p = .106$) comparisons were not significant after correction. These findings show that specifying a candidate's gender identity influenced judgments even when the professional narrative remained unchanged. Therefore, \textbf{H1 is fully supported that assigning a candidate gender identity changed the probability of a Yes vote when the remaining narrative information was unchanged}.

\begin{figure}[t]
\centering
\includegraphics[width=\linewidth]{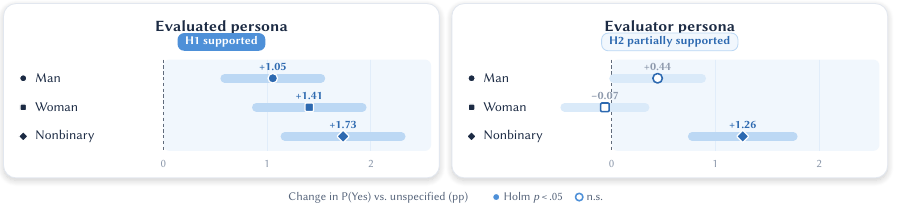}
\caption{Change in Yes-vote probability when an explicit gender label replaced the unspecified condition. Every explicit candidate gender identity increased Yes votes (1.05--1.73 percentage points, all adjusted $p < .001$), whereas only the nonbinary evaluator gender identity did so (1.26 points); the man and woman evaluator gender identities were not significant after Holm adjustment. Points show within-narrative estimates and bars show 95\% confidence intervals; filled markers indicate Holm-adjusted $p < .05$.}
\label{fig:gender-contrasts}
\end{figure}

\paragraph{Evaluator gender identity (H2).}
Evaluator gender assignments showed a more selective pattern. Relative to the unspecified condition, assigning a nonbinary evaluator identity increased Yes-vote probability by 1.26 percentage points (95\% CI [0.78, 1.75]), with an adjusted odds ratio of 1.13 (95\% CI [1.08, 1.18], Holm-adjusted $p < .001$). The 0.44-point increase for a man evaluator identity did not remain significant after Holm correction (95\% CI [0.02, 0.87], adjusted $p = .082$), while the difference for a woman evaluator identity was negligible ($-0.07$ points, 95\% CI [$-0.45$, 0.32], $p = .742$). Figure~\ref{fig:gender-contrasts} presents candidate and evaluator gender effects on the same scale. Secondary pairwise comparisons among explicit evaluator identities showed that nonbinary assignments increased Yes-vote probability relative to man (0.82 percentage points, Holm-adjusted $p < .001$) and woman (1.33 points, adjusted $p < .001$), while woman assignments decreased it relative to man ($-0.51$ points, adjusted $p = .004$). These findings demonstrate that the evaluator's assigned gender identity can influence judgments, including differences between explicitly stated identities. Therefore, \textbf{H2 was partially supported: assigning an evaluator gender identity significantly changed Yes-vote probability relative to the unspecified condition only for the non-binary assignment}.

\begin{table}[t]
\caption{Gender contrasts relative to the unspecified condition. All three candidate gender identity contrasts were significant (1.05--1.73 percentage points; odds ratios 1.10--1.17), whereas among evaluator gender identities only nonbinary was (1.26 points; odds ratio 1.13). The evaluator-man contrast did not remain significant after Holm adjustment.}
\label{tab:gender-results}
\centering\small
\begin{tabularx}{\linewidth}{@{}p{0.20\linewidth}p{0.17\linewidth}p{0.29\linewidth}X@{}}
\toprule
Factor & Label & $\Delta$ Yes (pp), 95\% CI & $\mathit{OR}$ [95\% CI] \\
\midrule
Evaluated & Man & 1.05 [0.59, 1.52] & 1.10 [1.06, 1.15] \\
 & Woman & 1.41 [0.90, 1.92] & 1.14 [1.09, 1.19] \\
 & Nonbinary & 1.73 [1.17, 2.29] & 1.17 [1.12, 1.23] \\
\addlinespace
Evaluator & Man & 0.44 [0.02, 0.87] & 1.04 [1.00, 1.08] \\
 & Woman & $-0.07$ [$-0.45$, 0.32] & 0.99 [0.96, 1.03] \\
 & Nonbinary & 1.26 [0.78, 1.75] & 1.13 [1.08, 1.18] \\
\bottomrule
\end{tabularx}
\par\vspace{2pt}\raggedright\scriptsize
\textit{Note.} pp = percentage points. All candidate gender identity contrasts and the nonbinary evaluator contrast have Holm-adjusted $p < .001$. For the evaluator-man contrast, adjusted $p = .082$ for the paired probability difference and $.072$ for the adjusted odds ratio. The evaluator-woman paired contrast has $p = .742$. Probability differences average over the other balanced conditions. Odds ratios adjust for the other gender factor and evaluation model.
\end{table}

\paragraph{Identity matching (H3).}
We found no statistically significant overall association between matching candidate and evaluator gender identities and Yes-vote probability after accounting for their separate effects. Among the 17,280 judgments with explicit identities on both sides, Yes-vote rates were 88.1\% for matching identities and 87.9\% for nonmatching identities. The adjusted matching association was small ($\mathit{OR} = 1.02$, 95\% CI [0.99, 1.05], $p = .297$), equivalent to a 0.17-percentage-point difference in predicted Yes-vote probability. The candidate-by-evaluator interaction was also not significant (Wald $\chi^2(4) = 5.93$, $p = .204$; man, woman, and nonbinary $\mathit{OR}$s $= 0.99$, $1.05$, and $1.01$). These results indicate that the effects of the two assigned identities did not produce an overall preference for candidates sharing the evaluator's identity. Therefore, \textbf{Therefore, H3 was \underline{not} supported: we \underline{did not} detect a statistically significant overall association between matching candidate and evaluator gender identities and Yes-vote probability beyond their separate effects}.

\paragraph{Changes to individual voting decisions.}
Most votes remained unchanged across identity conditions, but some individual decisions differed even though the candidate's professional narrative was identical. Relative to the condition in which both identities were unspecified, 3.02\% of votes reversed when only candidate gender identity was assigned (95\% CI [2.24, 3.80]), 2.48\% when only evaluator gender identity was assigned (95\% CI [1.84, 3.18]), and 3.27\% when both identities were assigned (95\% CI [2.48, 4.11]). Most reversals were from No to Yes. Across the 15 alternative identity conditions, 8.28\% of narrative--model pairs reversed at least once (95\% CI [6.51, 10.16]), and at least one model reversed a vote for 32.5\% of narratives. In a limited repeatability check, 768 requests from 24 narratives were repeated three times, and 1.04\% of cells showed disagreement (95\% CI [0.13, 2.47]).

\paragraph{Variation across sampled narratives.}
Increases associated with explicit candidate gender identities appeared across the sampled age, occupation, and source-dataset groups, with no statistically significant differences in their magnitude. Candidate gender identity interactions were not significant for age ($\chi^2(3) = 3.45$, $p = .327$), occupation ($\chi^2(3) = 0.18$, $p = .980$), or source dataset ($\chi^2(3) = 3.05$, $p = .384$). Evaluator gender identity interactions were also not significant (all $p$s $> .34$). These secondary analyses concern the sampled narratives because age and occupation were not manipulated, and occupation categories differed between the US and French datasets. 

\subsection{RQ2: How Do Voting Decisions and Gender-Identity Effects Differ Between LaJ and DLMs?}

\begin{figure}[ht]
\centering
\includegraphics[width=0.7\linewidth]{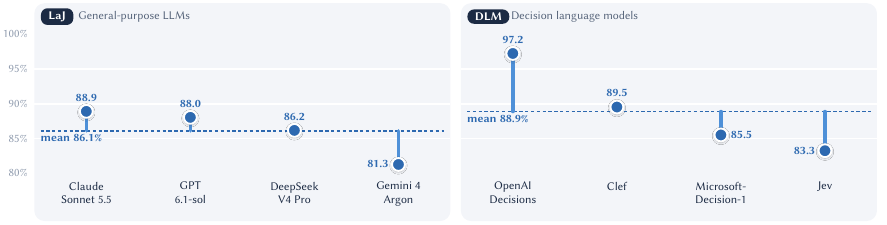}
\caption{Yes-vote rates for the eight evaluation models. Rates ranged from 81.3\% for Gemini 4 Argon to 97.2\% for OpenAI Decisions, a spread larger than any gender effect. DLMs returned Yes more often than LaJ models on average (88.9\% vs.\ 86.1\%), but rates varied widely within both groups. Each line runs from the group mean (dashed) to the model's rate; each model contributed 3,840 judgments.}
\label{fig:model-rates}
\end{figure}

\paragraph{Differences in voting decisions (H4).}
Yes-vote rates differed substantially across evaluation models, exceeding the average differences associated with explicit gender assignments. Rates ranged from 81.3\% for Gemini 4 Argon to 97.2\% for OpenAI Decisions, a difference of 15.9 percentage points (Figure~\ref{fig:model-rates}). The overall model comparison was significant (Wald $\chi^2(7) = 65.11$, $p < .001$). DLMs also returned Yes votes more frequently than LaJ models (88.9\% vs. 86.1\%), a paired difference of 2.79 percentage points (95\% CI [1.50, 4.07], $p < .001$). The adjusted category contrast was significant ($\mathit{OR} = 1.29$, 95\% CI [1.16, 1.44], $p < .001$). These results demonstrate differences in voting outcomes both among individual evaluation models and between the two model categories. Therefore, \textbf{H4 is fully supported as the probability of a Yes vote differed across evaluation models under equivalent conditions}.

\begin{figure}[ht]
\centering
\includegraphics[width=0.7\linewidth]{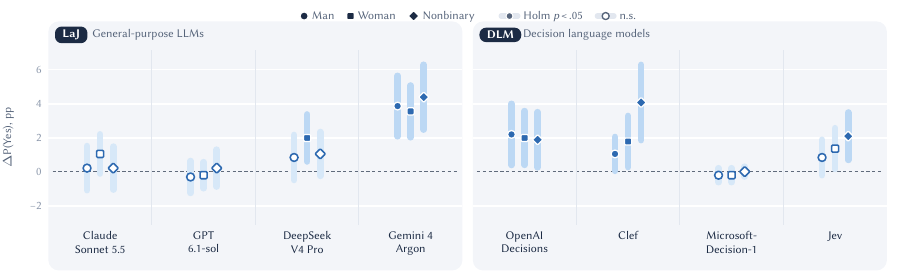}
\caption{Candidate gender identity effects by evaluation model. Gemini 4 Argon showed the largest effects, with all three labels increasing Yes votes by 3.54--4.38 percentage points, whereas no label had a significant effect for Claude Sonnet 5.5, GPT-6.1-sol, or Microsoft-Decision-1. Estimates are changes in Yes-vote probability relative to the unspecified candidate gender identity; bars show 95\% confidence intervals, marker shape indicates gender, and filled markers indicate Holm-adjusted $p < .05$ within the model.}
\label{fig:candidate-by-model}
\end{figure}

\begin{figure}[ht]
\centering
\includegraphics[width=0.7\linewidth]{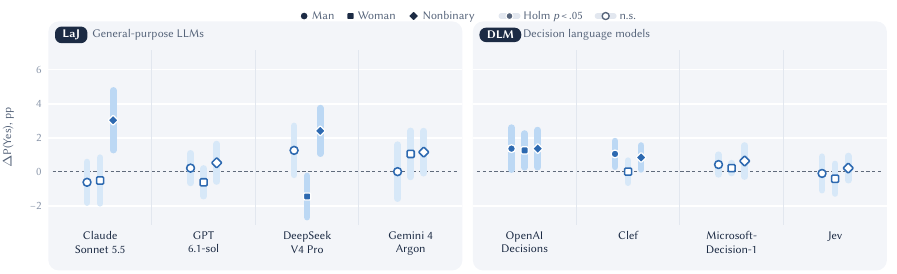}
\caption{Evaluator gender identity effects by evaluation model. The pattern differed from the candidate gender identity effects: a nonbinary evaluator gender identity increased Yes votes for Claude Sonnet 5.5 (3.02 percentage points) and DeepSeek V4 Pro (2.40), while no evaluator gender identity had a significant effect for Gemini 4 Argon. DeepSeek V4 Pro also showed the only significant negative contrast ($-1.46$ points for a woman evaluator gender identity). Estimates are relative to the unspecified evaluator gender identity; encoding as in Figure~\ref{fig:candidate-by-model}.}
\label{fig:evaluator-by-model}
\end{figure}

\paragraph{Differences in gender sensitivity (H5).}
The effects of candidate and evaluator gender assignments also varied across evaluation models. Candidate gender contrasts reached 3.54--4.38 percentage points for Gemini 4 Argon, while none of the corresponding contrasts remained significant after Holm correction for Claude Sonnet 5.5, GPT-6.1-sol, or Microsoft-Decision-1 (Figure~\ref{fig:candidate-by-model}). Evaluator gender effects followed a different pattern: assigning a nonbinary evaluator identity increased Yes-vote probability by 3.02 points for Claude Sonnet 5.5 and 2.40 points for DeepSeek V4 Pro, while assigning a woman evaluator identity decreased it by 1.46 points for DeepSeek V4 Pro (Figure~\ref{fig:evaluator-by-model}). The overall model-by-gender interaction was significant (Wald $\chi^2(42) = 102.30$, $p < .001$). Matching associations also varied across models (Wald $\chi^2(7) = 15.97$, $p = .025$), although the pooled matching association was not significant. These results show that sensitivity to assigned identities depended on the individual evaluation model. Therefore, \textbf{H5 is fully supported, thus the effects of candidate gender identity, evaluator gender identity, and identity matching differed across evaluation models}.

At the category level, the overall gender-by-category interaction was significant (Wald $\chi^2(6) = 20.08$, $p = .003$). However, no individual category difference for a specific gender contrast remained significant after Holm correction, and the identity-matching association did not differ significantly between LaJ and DLMs (Wald $\chi^2(1) = 1.41$, $p = .235$). The overall category difference in voting rates therefore did not translate into consistent differences in sensitivity to specific gender assignments.

\begin{figure*}[t]
\centering
\includegraphics[width=0.7\textwidth]{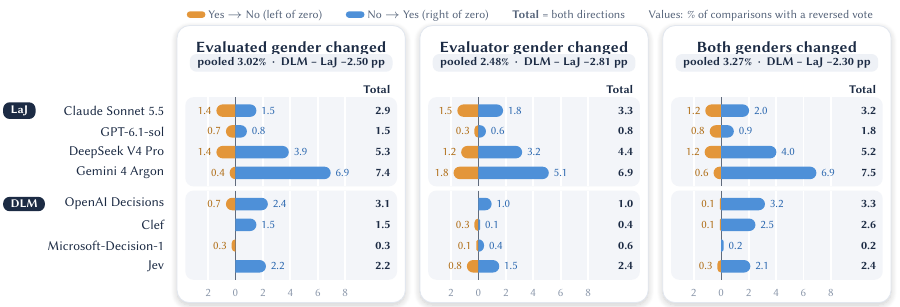}
\caption{Decision reversals relative to the unspecified/unspecified baseline for the same narrative and evaluation model. Most reversals were from No to Yes. Gemini 4 Argon reversed 7.36\% of its votes when only the candidate gender identity changed, compared with 0.28\% for Microsoft-Decision-1, and DLMs reversed less often than LaJ models under all three manipulations ($-2.30$ to $-2.81$ percentage points). Bars left of zero show Yes-to-No changes and bars right of zero show No-to-Yes changes; panel headers give the pooled reversal rate and the DLM--LaJ difference.}
\label{fig:decision-reversals}
\end{figure*}

\paragraph{Model differences in decision reversals.}
Decision reversals also varied across individual models and model categories. When only candidate gender identity was assigned, Gemini 4 Argon reversed 7.36\% of votes and DeepSeek V4 Pro reversed 5.28\%, compared with 0.28\% for Microsoft-Decision-1 (Figure~\ref{fig:decision-reversals}). DLM reversal rates were lower than LaJ rates by 2.50, 2.81, and 2.30 percentage points for candidate-only, evaluator-only, and joint identity assignments, respectively (all Holm-adjusted $p$s $< .001$). These differences indicate greater observed decision stability among the DLMs tested, but they do not establish greater fairness or decision correctness.

\paragraph{Gender sensitivity under explicit identity substitutions.}
Secondary analyses compared decisions when one explicit gender identity was replaced with another while the professional narrative and the other identity assignment remained unchanged. Across 46,080 paired comparisons, 2.28\% of votes differed between gender assignments. DLMs exhibited fewer reversals than LaJ models for both candidate substitutions (1.35\% vs.\ 3.07\%; difference $-1.72$ percentage points, 95\% CI [$-2.62$, $-0.87$]) and evaluator substitutions (1.01\% vs.\ 3.68\%; difference $-2.67$ points, 95\% CI [$-3.70$, $-1.65$]). Across both manipulations, DLMs reversed 1.18\% of decisions compared with 3.38\% for LaJ models, a difference of $-2.20$ percentage points (95\% CI [$-3.13$, $-1.30$], bootstrap $p < .001$). All six gender-pair category differences remained significant after Holm correction. The overall difference persisted in leave-one-model-out analyses, although a model-level permutation test was not significant ($p = .086$). These findings indicate lower observed decision-level gender sensitivity among the DLMs tested, with differences across individual models.

A secondary country-level analysis indicated that the difference between model categories also varied by narrative country. In the US sample, DLMs and LaJ models returned Yes votes in 90.7\% and 86.5\% of cases, respectively, compared with 87.1\% and 85.7\% in the French sample. The category-by-source interaction was significant (Wald $\chi^2(1) = 7.48$, $p = .006$). This difference should not be interpreted as a country effect because the samples differed in occupations, narrative content, and potentially language. Additional estimates are provided in the supplementary analysis.

\section{Discussion}

\subsection{Theoretical Implications}

First, our study extends research on gender sensitivity in AI evaluations by examining the identities of both candidates and evaluators within the same judgment. Previous studies have examined gender-related differences in candidate assessments \cite{an2024hiring,seshadri2025hiring}, while research on persona prompting has shown that assigned demographic identities can influence model responses \cite{neumann2025position,lutz2025prompt}. We connect these streams by distinguishing three sources of variation: disclosing a candidate's gender, substituting one explicit gender identity for another, and assigning a gender identity to the evaluator. Explicit candidate identities increased favorable judgments relative to unspecified gender, and direct comparisons showed differences among the stated identities. Evaluator assignments followed a different pattern, including differences between explicitly assigned evaluator identities. Yet matching candidate and evaluator identities showed no significant overall association beyond their separate effects. This qualifies expectations derived from studies of conformity among persona-conditioned agents \cite{li2026societal}: shared identities in an interaction do not necessarily produce more favorable evaluations. Conceptually, the identity assigned to an AI evaluator is itself part of the evaluation conditions, even though it provides no additional evidence about the candidate. Gender sensitivity therefore encompasses distinct candidate, evaluator, and relational effects; observing one does not establish the presence of another.

Second, our comparison extends research on model-dependent demographic sensitivity \cite{brown2026context,ye2025justice} to decision language models. When one explicit gender identity was substituted for another, the four DLMs reversed 1.18\% of decisions compared with 3.38\% for the four LaJ models, approximately 65\% fewer observed reversals. Lower DLM reversal rates appeared for both candidate and evaluator substitutions, across all six gender-pair contrasts, and remained evident when individual models were excluded. This supports a specific conclusion: the DLMs tested exhibited lower \emph{decision-level gender sensitivity} under controlled identity substitutions. It does not imply that they were insensitive to gender or that they made fairer decisions. Individual models differed substantially, and an exact model-level permutation comparison did not establish a general difference between the broader LaJ and DLM categories ($p = .086$). Moreover, model identity and request interface varied together, preventing attribution of the difference to structured decision interfaces alone. These results show why demographic sensitivity must be examined at both the model and evaluation-approach levels, with attention to the particular outcome used to measure it.

Third, our findings distinguish aggregate gender effects from sensitivity at the level of individual decisions. Previous research often compares demographic groups using average scores, acceptance rates, or rankings \cite{gaebler2025auditing,wang2024jobfair}, while counterfactual audits also examine whether decisions change when demographic information changes \cite{zenginoglu2026credit}. Our results show that relatively small differences in overall Yes-vote probability can coexist with reversals for an individual candidate whose professional information remains unchanged. Aggregate differences describe systematic shifts in favorable judgments, whereas reversal rates capture changes in the decisions returned for the same candidate under controlled identity assignments. This distinction is consequential for binary evaluations: a small shift in an overall rate can still correspond to a different outcome for particular individuals. Reversal direction is also informative because movement from No to Yes and from Yes to No may be obscured by a single net difference. Neither measure, however, determines whether the original or changed decision was justified. Consistency, demographic sensitivity, and fairness are related but distinct properties, and conclusions about one should not be substituted for evidence about another.

\subsection{Practical Implications}

These findings have implications for institutions considering AI-assisted evaluations of people, although our task did not involve a real electoral process. We propose five practical guidelines (GLs):

\begin{itemize}
    \item \textbf{GL1: Test both identities.} Vary candidate and evaluator gender assignments independently while holding professional information and evaluation criteria fixed. Include unspecified and explicitly assigned identities.

    \item \textbf{GL2: Examine identity matching.} Compare matching and nonmatching identities after accounting for the separate effects of each identity, rather than treating agreement in identities as evidence of favoritism.

    \item \textbf{GL3: Measure decision reversals.} Report Yes-to-No and No-to-Yes changes under identity disclosure and explicit gender substitutions alongside average voting differences.

    \item \textbf{GL4: Compare evaluation models.} Evaluate LaJ and DLM systems on the same cases and report both category-level and model-specific sensitivity. Lower reversal rates should not be interpreted as greater fairness.

    \item \textbf{GL5: Assess decision quality and accountability.} Review whether judgments follow stated criteria, document the basis of decisions, and provide opportunities to question outcomes that affect people.
\end{itemize}

\subsection{Limitations and Future Work}

Five limitations qualify our conclusions. \textit{First}, the synthetic Nemotron-Personas narratives covered a limited range of occupations and age groups. The US and French datasets were not matched, and implicit gender cues may have remained despite preprocessing. \textit{Second}, the hypothetical city council task lacked human evaluators and an independent benchmark of candidate suitability. Its high overall Yes-vote rate (87.5\%) also constrains the observable pattern of reversals, limiting conclusions about practical decision quality. \textit{Third}, we could not verify whether models adopted the assigned evaluator identities or inferred gender when no identity was specified. \textit{Fourth}, only four models represented each evaluation category, and model differences were confounded with interface differences. The observed DLM advantage therefore does not establish an inherent advantage of DLMs as a class. \textit{Fifth}, repeated identical-input evaluations covered only 24 narratives; ordinary run-to-run variation cannot be completely separated from the effects of identity substitutions.

Future research could replicate these comparisons with more diverse narratives and demographic attributes, include human evaluators and independently assessed candidate suitability, and examine how identity instructions are interpreted. Comparing conversational and structured decision interfaces for the same underlying models would help isolate interface effects. Larger repeated-measures studies could also clarify the stability of individual decisions and whether lower gender-substitution reversal rates persist across tasks and candidate populations. Such extensions would establish how far the present findings generalize and whether differences in consistency correspond to differences in evaluation quality.

\section{Conclusion}

We examined how candidate gender identities, evaluator personas, and model choice shape AI-assisted voting decisions. Both candidate and evaluator identities affected judgments, while identity matching showed no significant overall association. Under explicit gender substitutions, the tested DLMs exhibited fewer decision reversals than LaJ models, alongside substantial model-specific variation. By separating identity disclosure, identity substitution, and matching, and by comparing aggregate outcome differences with individual reversals, our findings offer a more precise account of gender sensitivity in AI evaluation. Audits should examine these dimensions together while keeping decision consistency distinct from fairness and correctness.

\bibliographystyle{ACM-Reference-Format}
\bibliography{references}
\end{document}